\documentclass{article}
\usepackage{spconf,amsmath,epsfig}
\usepackage{color}
\usepackage{amsfonts,amssymb}
\usepackage{algorithm}
\usepackage{algorithmic}
\usepackage{amsmath}
\usepackage{multirow}
\usepackage{hhline}
\usepackage{colortbl}
\usepackage{bbding}
\usepackage{pifont}
\usepackage{makecell}
\usepackage{booktabs}
\usepackage{microtype}
\usepackage{amsfonts}

\usepackage[pagebackref,breaklinks,colorlinks]{hyperref}
\usepackage[capitalize]{cleveref}
\crefname{section}{Sec.}{Secs.}
\Crefname{section}{Section}{Sections}
\Crefname{table}{Table}{Tables}
\crefname{table}{Tab.}{Tabs.}

\usepackage{xcolor}         % colors
\usepackage{newfloat}
\usepackage{listings}
\usepackage{authblk}
\usepackage{changes}
\usepackage{lipsum}% <- For dummy text
\definechangesauthor[name={Yuchen.Ji}, color=blue]{yuchen}

\let\OLDthebibliography\thebibliography
\renewcommand\thebibliography[1]{
  \OLDthebibliography{#1}
  \setlength{\parskip}{0pt}
  \setlength{\itemsep}{0pt plus 0.3ex}
}

\begin{document}\sloppy

% Example definitions.
% --------------------
\def\x{{\mathbf x}}
\def\L{{\cal L}}

% Title.
% ------
% \title{Hierachical Refinement for Indoor Scene Reconstruction with rich Details}
\title{Fine-detailed Neural Indoor Scene Reconstruction using multi-level importance sampling and multi-view consistency}
% \title{Fine-detailed neural implicit surface reconstruction for indoor scene}
%
% Single address.
% ---------------
\name{Author(s) Name(s)\thanks{Thanks to XYZ agency for funding.}}
% \address{Author Affiliation(s)}

% \twoauthors{Xinghui Li}{Tsinghua University}{author}{Huawei}
\author[1,3]{Xinghui Li}
\author[2]{Yuchen Ji}
\author[1]{Xiansong Lai}
\author[1]{Wanting Zhang}
\author[1$*$]{Long Zeng\thanks{\textsuperscript{$*$}Corresponding author}}
\affil[1]{Tsinghua University, China}
\affil[2]{Nanjing University of Aeronautics and Astronautics, China}
\affil[3]{ByteDance, China}
% \renewcommand*{\Authsep}{   }
% \renewcommand*{\Authands}{   }
% \date{}
\address{}
\maketitle

%
% For example:
% ------------
%\address{School\\
%	Department\\
%	Address}
%
% Two addresses (uncomment and modify for two-address case).
% ----------------------------------------------------------
%\twoauthors
%  {A. Author-one, B. Author-two\sthanks{Thanks to XYZ agency for funding.}}
%	{School A-B\\
%	Department A-B\\
%	Address A-B}
%  {C. Author-three, D. Author-four\sthanks{The fourth author performed the work
%	while at ...}}
%	{School C-D\\
%	Department C-D\\
%	Address C-D}
%
% \begin{document}
%\ninept
%
% \maketitle
%
\begin{abstract}
Recently, neural implicit 3D reconstruction in indoor scenarios has become popular due to its simplicity and impressive performance. Previous works could produce complete results leveraging monocular priors of normal or depth. However, they may suffer from over-smoothed reconstructions and long-time optimization due to unbiased sampling and inaccurate monocular priors. In this paper, we propose a novel neural implicit surface reconstruction method, named FD-NeuS, to learn fine-detailed 3D models using multi-level importance sampling strategy and multi-view consistency methodology. Specifically, we leverage segmentation priors to guide region-based ray sampling, and use piecewise exponential functions as weights to pilot 3D points sampling along the rays, ensuring more attention on important regions. 
% In addition, We introduce multi-view feature consistency as supervision to improve the reconstruction of details. Furthermore, our method utilizes multi-view normal consistency as uncertainty to filter inaccurate normal priors and further to guide ray sampling. 
In addition, we introduce multi-view feature consistency and multi-view normal consistency as supervision and uncertainty respectively, which further improve the reconstruction of details. Extensive quantitative and qualitative results show that FD-NeuS outperforms existing methods in various scenes. 
% across various evaluation metrics.
\end{abstract}
\section{Introduction}
\label{sec:intro}
3D scene reconstruction from multi-view RGB images is an important and challenging task in computer vision. Traditional methods~\cite{wang2018mvdepthnet,long2020occlusion} estimate dense depth maps for each image and then fuse them to 3D models. Such methods often get noisy surfaces and incomplete geometry due to inconsistent predictions at each frame. Some other methods, such as volumetric methods ~\cite{murez2020atlas,sun2021neuralrecon} use explicit voxels to model 3D scenes and directly regress input images to TSDF value or sparse occupancy. Although such methods yield better completeness, their limited voxel resolutions result in poor details.

Recently, NeRF-based methods~\cite{mildenhall2021nerf,wang2021neus,yariv2021volume} have attracted increasing attention due to impressive reconstruction results, which model the scenes using neural implicit representations. ~\cite{wang2021neus,yariv2021volume} represent the geometry of 3D scenes as signed distance fields (SDF), improving geometry quality. 
% They can recovery high-quality object-level surfaces from images.
However, these methods essentially rely on multi-view photometric consistency to learn implicit representations, leading to poor performance in texture-less regions indoors. To address the challenge,~\cite{Guo_2022_CVPR,yu2022monosdf} leverage priors about the indoor scenes, such as Manhattan world assumption~\cite{Guo_2022_CVPR} and pseudo depth supervision~\cite{yu2022monosdf}. ~\cite{yu2022monosdf,wang2022neuris} further improve the reconstruction quality by adopting monocular normal priors based on the observation of great planarity in textureless regions indoors. While the primary structure, such as walls and floors of indoor scene, can be reasonably reconstructed, these methods still struggle to recover fine details due to the low sampling probabilities of detailed regions and inaccurate monocular priors. 
% , leading to overly-smoothed in non planar area.
% Although [NeuRIS] judges normal uncertainty based on multi-view photometric consistency, the problem still exists. 

In this work, we present a novel neural surface reconstruction method named \textbf{FD-NeuS}, aiming to address the problems of missing details and over-smoothed reconstruction in indoor scenes. 
% \replaced{Existing pixel sampling and point sampling strategies along rays do not sample enough detail areas, resulting in a smooth trend with surrounding areas, making it difficult to maintain the integrity of details.}{
Due to the observation of low sampling probability in detail areas and inefficient points sampling around the surface in the original Hierarchical Volume Sampling (HVS) strategy, we propose a multi-level importance sampling strategy to improve the efficiency and accuracy of the sampling phase. Specifically, we first use a segmentation detection network to obtain the segmentation map with fine mask for each image and then use our region-based ray importance sampling strategy to train the neural implicit network, which not only provides more attention to the challenging detailed areas, but also improves the sampling efficiency compared with traditional random sampling. At the same time, we utilize piecewise exponential functions rather than original constant functions as Probability Density Function (PDF), to guide points sampling along the ray, which enables the sampling points to approximate the potential surface more quickly and accurately. In addition, we add multi-view feature consistency at only the surface point along the sampling ray to further improve local geometric details. Furthermore, we use multi-view normal consistency to filter unreliable normal priors and increase the sampling of unreliable areas. As a result, extensive experiments in various scenes show that FD-NeuS achieves start-of-art performance in reconstructing indoor scenarios. 
% 由于现有的像素采样和沿射线的点采样策略，对于细节区域的采样量不足，导致其与周围区域呈现平滑趋势，从而难以保持细节特征的完整性。

% \added{Sufficient ablation studies also prove the effectiveness of the each improvements in our method.}{}

% Specially, FD-NeuS includes two main aspects of improvement, multi-level resampling and multi-view consistency. For multi-view resampling, we first use SAM and a classifier to obtain the semantic map for each image and then use power functions to guide ray sampling in adaptive manner, which not only provides more attention in the challenging detailed area, but also improves the utilization efficiency of sampling data. For multi-view consistency, we add multi-view feature consistency at only the surface points along the sampling ray, to improve the local geometric details. Futhermore, we use multi-view normal consistency to filter unreliable normal priors and further to increase the sampling of unreliable areas. As a result, extensive experiments in various scenes show that FD-NeuS achieves start-of-art performance in reconstructing indoor scenarios. Sufficient ablation studies also prove the effectiveness of the each improvements in our method.

In summary, our contributions are as follows:

\begin{itemize}
  \item We propose FD-NeuS, a novel neural implicit surface reconstruction method that can recover fine details for complex indoor scenes.

  \item We introduce a multi-level importance sampling methodology, including ray level and point level, to offer more attention to detailed regions and potential surfaces respectively, which also improves the efficiency of sampling phase.
  % edge-guided sampling strategy to address the problems of detail region missing reconstruction and low efficiency caused by traditional random sampling.

  % \item We conditionally adopt the normal prior, which is achieved by judging whether the location is near to edge areas. The strategy enables reasonable reconstruction of geometric details with sharp shapes.

  \item We introduce multi-view consistency as supervision and uncertainty to guide the optimization, further improving the reconstruction quality of details.
  
  % \item We apply additional supervision of SfM sparse points, making the reconstruction of regions with rich visual features more complete and accurate.
  % \item Finally, for points sampling along the ray, we adopt an end-point guided sampling strategy based on ray casting, and dynamically update its boundary during training to make point sampling more efficient.

  \item Extensive experiments on various scenes show that our method achieves SOTA performance in multiple metrics.
\end{itemize}

% contribution是否可以按照这样的方式描述：
% ===== NeuRIS =====
% 1. 写提出了这个方法，然后是这个方法的优势，用来解决什么问题
% 2. 写改进点一
% 3. 写改进点二
% 最后用一段来描述额外的实验结果
% ===== NeuralRoom =====
% 将In summary, our contributions are as follows换成，To summarize, FD-NeuS has the following advantages:
% 2. 写改进点一
% 3. 写改进点二
% 最后用一段来描述额外的实验结果

% We propose FD-NeuS, a novel neural implicit surface reconstruction method, (which uses multi-level resampling and multi-view consistency) to improve the reconstruction quality of detailed areas, while improving data utilization effciency without extra training time.
% We introduce multi-level resampling to guide ray sampling and points sampling through semantic and geometric priors, making sampling points pay more attention to details and surface, so that improve utilization efficiency and accuracy of sampling points.
% We introduce multi-view consistency as supervision and uncertainty to guide the optimization, which further improve the reconstruction quality of detailed regions. (还是你写的合适啊)
% Experimental results on xxx datasets/x scenes show that FD-NeuS achieve superior reconstruction result in indoor scene, especially in fine details. Ablation studies demonstrate the effectiveness of multi-level resampling and multi-view consistency.

% 我们引入了muli-level resampling，通过语义和几何先验引导ray sampling和points sampling，使采样点更关注细节和物体表面，从而提高采样点的利用效率和准确性。
% xxxxxxx我们引入了multi-view consistency, 利用多视图的特征一致性，优化对细节区域的重建，利用多视图的normal一致性来过滤不准确的法向先验，并引导对不准确区域的采样偏置。

\section{Related Work}\label{sec:format}
\noindent\textbf{MVS-based Explicit Reconstruction}
% Traditional multi-view stereo methods~\cite{schonberger2016structure} achieve great success in textured surfaces, but tend to show poor reconstruction results in texture-less regions. Recently, some works leverage deep learning techniques to overcome this problem.
% Learning-based 
% \added{(Is 'Per-view depth estimation-based' necessary?)}{}
Per-view depth estimation-based multi-view stereo (MVS) methods~\cite{wang2018mvdepthnet,long2020occlusion} predict depth map for each image and fuse them to form a point cloud, which can be subsequently processed by using meshing algorithms~\cite{lorensen1987marching} to get complete surface. However, these methods suffer from redundant computation and poor consistency. In this work, we learn a coordinate-based implicit neural scene representation rather than fusing depth maps from multi-views.

\noindent\textbf{Neural Scene Reconstruction}
Neural scene reconstruction models the properties (e.g., occupancy, TSDF) of 3D positions using neural networks. \cite{murez2020atlas} first proposes a volumetric design, which uses voxels that store TSDF value as the representation of scenes.
% extracts features from the input images, unprojects the features into 3D space, and uses 3D CNN to regularize the features to predict the TSDF value.
\cite{sun2021neuralrecon} divides the space into multiple fragments and utilizes a recurrent network to fuse the features from previous fragments sequentially.
% They suffer from high memory consumption due to representing scenes as discrete voxels.
Recently, \cite{liu2020dist, yariv2020multiview} used a coordinate-based implicit neural function to model the 3D space and show impressive performance in reconstruction. 
% Recently, neural implicit scene representations show impressive performance in indoor reconstruction, which use implicit functions to model the 3D space that is conditioned on 3D coordinates \cite{liu2020dist,yariv2020multiview}.
% have become a popular way to represent 3D geometry for their expressiveness and low memory consumption. 
% \cite{liu2020dist,yariv2020multiview} represent surfaces via a single MLP based on implicit differentiable rendering and result in continuous outputs.
% Although they achieve impressive reconstruction, accurate object masks are required.
Inspired by the success of NeRF~\cite{mildenhall2021nerf}, NeuS~\cite{wang2021neus} and VolSDF~\cite{yariv2021volume} transform SDF to volume density and use volume rendering for neural implicit surface reconstruction. However, these methods show poor performance in texture-less planar regions. In this work, we incorporate additional monocular priors to guide the geometry optimization process, which recovers fine details in challenging indoor scenes.

% \noindent\textbf{Geometry Priors for NeRF}
% Several works have proposed to introduce geometry priors while optimizing NeRF such as sparse SfM point clouds, semantic similarity~\cite{jain2021putting}, or dense depth information~\cite{roessle2022dense}. However, these works is designed for novel view synthesis. Recently, some researchers incorporate priors to the task of 3D surface reconstruction. Manhattan-SDF~\cite{Guo_2022_CVPR} uses Manhattan world priors to handle textureless regions including walls and floors, which is limited by the accuracy of semantic segmentation. Related and concurrent to our work are NeuRIS~\cite{wang2022neuris} and MonoSDF~\cite{yu2022monosdf}, which show the importance of normal prior to the reconstruction indoor. Although they achieve amazing reconstruction results in texture-less region, we experimentally find that these methods tend to cause missing reconstruction and overly smooth reconstruction for detailed areas of indoor scenes. 
% Thus, we additionally adopt the segmentation prior to guide ray sampling, and leverage multi-view consistency, which in turn improve the reconstruction quality.

\section{Methodology}\label{sec:Method}
\begin{figure}[!t]
% \vspace{-0.1cm}
\centering
\includegraphics[scale=0.32]{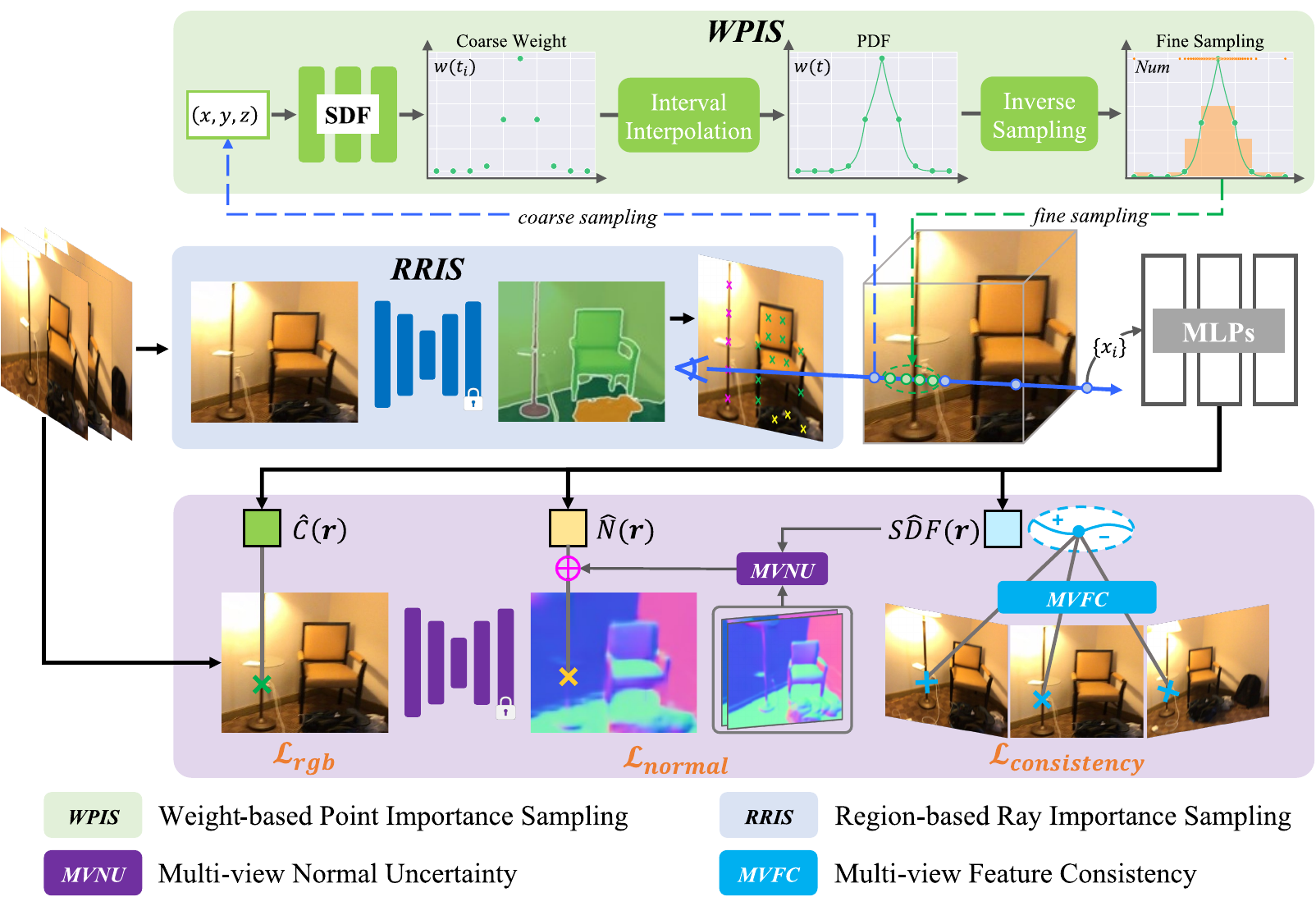}
\vspace{-0.3cm}
\caption{\textbf{Method overview of FD-NeuS}. We utilize segmentation priors to achieve region-based ray importance sampling and use piece-wise exponential functions as weights to guide point importance sampling. Additionally, we adopt multi-view feature consistency as supervision, and use multi-view normal consistency as uncertainty to filter unreliable normal priors.}
% \vspace{-0.9cm}
\label{Fig. overview}
\end{figure}

Given multi-view posed images, our goal is to accurately reconstruct fine-detailed scene geometry. To this end, we represent scene geometry and textures as signed distance functions and color fields, which utilize volume rendering technique to optimize (Sec. \ref{sec:Preliminary}). 
% To reconstruct high-quality indoor scenes not only containing large texture-less areas but also irregular shapes with fine details, 
To reconstruct fine-detailed indoor scenes, we propose a multi-level importance sampling strategy and adopt multi-view consistency methodology. 
Specifically, to ensure more attention on important areas, we utilize segmentation map to guide region-based ray importance sampling, and leverage piecewise exponential functions as PDF to pilot points sampling along the rays (Sec. \ref{sec:Multi-level-Resampling}). 
To further strengthen the learning of detailed regions, we introduce multi-view constraints, including feature consistency and normal consistency, which are respectively used to apply explicit supervision and perform as uncertainty to guide optimization (Sec. \ref{sec:Multi-view Consistency}). Finally, we discuss the loss functions and the overall optimization process (Sec. \ref{sec:Training}). ~\cref{Fig. overview} shows the overview of our method.

%%%%%%%%%%%%%%%%%%%%%%%%%%%%%%%%%%%%%%%%%%%%%%%%%%%%%%%
\linespread{1.0}
%%%%%%%%%%%%%%%%%%%%%%%%%%%%%%%%%%%%%%%%%%%%%%%%%%%%%%%

\subsection{Preliminary}\label{sec:Preliminary}
Following~\cite{wang2021neus}, we represent an indoor scene using two multilayer perceptrons (MLPs): geometry network $ f_g \colon \mathbb R ^3 \rightarrow \mathbb R$ maps a spatial position $\mathbf{x}\in \mathbb{R}^3$ to the signed distance function (SDF), and color network $f_c\colon\mathbb R^3\times \mathbb S^2\rightarrow \mathbb R^3$ maps $\mathbf x\in \mathbb R^3$ and a view direction $\mathbf v\in \mathbb S^2$ to the color. The surface $S$ of scene geometry is presented by the zero level-set of the SDF $S=\{\mathbf x \in \mathbb R^3 \mid f_g(\mathbf x)=0\}$.
% \begin{equation}
% S=\{\mathbf x \in \mathbb R^3 \mid f(\mathbf x)=0\}.
% \end{equation}

To optimize the implicit representation based on the supervision of 2D image observations, we adopt the volume rendering methodology. 
% 具体的，对于一个像素p会生成一条穿过光心和像素点的射线r，射线r上的离散3D点r(t_i)=o+vt_i被f_g和f_c分别映射为符号距离场和颜色，对射线r上的点以tn和tf为边界定积分可以得到该像素的颜色：
% 射线上的离散点t的3D坐标
Specially, for pixel $p$, we sample $N$ points $\left\{t_i|i=1,...,N\right\}$ along the ray $\mathbf{r}$. These points' 3D coordinate $\{\mathbf{x}(t_i)=\mathbf{o}+t_i\mathbf{v}\}_{i=1}^{N}$ are mapped into SDF and color using $f_g(\cdot)$ and $f_c(\cdot)$ respectively, where $\mathbf{o}$ is camera center. Therefore, the color of pixel $p$ can be obtained by numerically integrating the SDF and color of points along the ray $\mathbf{r}$:

\begin{equation}
    \Hat{C}(p)=\sum_{i=1}^{N}\omega(t_i)f_c(t_i),
\end{equation}

\begin{equation}
    \omega(t_i)=\sum_{i=1}^{N}T_i\alpha_i,
\end{equation}

\begin{equation}
    \alpha_i=1-\exp\Big (-\int^{t_{i+1}}_{t_i}\rho(t) {\rm d}t\Big ),
\end{equation}
where $T_i=\prod^{i-1}_{j=1}(1-\alpha_j)$ is the accumulated transmittance, $\omega(t_i)$ and $\alpha_i$ respectively represent the weight and discreate opacity of sample point $t_i$ along the ray $\mathbf r$, and $\rho(t)$ is the opaque density following the definition in NeuS~\cite{wang2021neus}.
Similarly, the normal can be rendered as $\hat{\mathcal{N}}(p)=\sum_{i=1}^N T_i\alpha_i\hat{\mathbf n}_i$,
% \begin{equation}
% \hat{\mathcal{N}}(\mathbf r)=\sum ^\mathit{n}_{i=1}T^i_{\mathbf r}\alpha^i_{\mathbf r}\hat{\mathbf n}^i_{\mathbf r},
% \end{equation}
where $\hat{\mathbf n}_i=\nabla f_n(t_i)$ denotes the derivative of SDF at point $t_i$, which can be calculated by PyTorch’s automatic derivation. 

% \subsection{ray resampling at image level}
\subsection{Multi-level Importance Sampling} \label{sec:Multi-level-Resampling}
% To reconstruct high-quality 3D models with fine details of indoor scenes, we propose a multi-level resampling strategy including ray level and points level. We first utilize semantic segmentation information, which plays a role in remapping sampling probabilities adaptively to guide efficient ray sampling. Additionally, we replace the traditional piecewise constant functions with piecewise exponential functions as Probability Density Function(PDF) to better sample points around the surface. % or potential surface?   可能有些冗长，看后续篇幅决定是否删减

To reconstruct high-quality indoor scenes with fine details, we propose a multi-level importance sampling strategy, including ray and point levels. This section describes how to use the strategy to guide sampling. % 精简版

% 首先引出随机采样的细节采样不到的问题，再提出有一些方法利用feature和edge做重要性采样但在一些case低效，从而引出我们方法的卓越性：利用类别占比面积去做动态概率重映射，同时保证场景学习过程中动态变化重要性的adaptivity
% 这部分需要再说一下利用semantic的信息对edge做了过滤进一步增强采样的准确性和额外利用语义分割的boundary作为edge指导采样吗?
\noindent\textbf{Region-based Ray Importance Sampling} 
Random sampling is a straightforward strategy and is widely used by NeRF-based works~\cite{wang2021neus,yu2022monosdf,wang2022neuris}, which uniformly selects $q$ rays from all pixels on the input image, leading to relative ignorance of the details in the corresponding scene. 
% However, compared with the textureless flat areas like walls and floors, detailed regions account for a small proportion, so random sampling is more likely to result in low sampling probabilities of these regions in the corresponding scene. 
Some works~\cite{li2023p, li2023edge} utilize image features or edge information to locate detailed areas and guide ray sampling. However, these methods do not generalize well since some planar surfaces also have rich texture features. 
% 细节区域重建不好主要是因为他们在场景所占的比例小从而导致被采样到的频次低，但重建这些拥有高频信息的区域本身就是困难的，因此我们提出了一种区域性采样从而显式的增加对占比小的物体的关注
Noticing that the main reason for details missing is the low sampling probabilities of detailed regions due to the small proportion compared with the flat areas like walls and floors. We propose a novel region-based ray importance sampling strategy to guarantee the rays of each partition can be sampled at each iteration, which keeps the balance between the textureless areas and the details in the training process. 
% We instead utilize segmentation priors to guide region-wise sampling, which not only explicitly pay attention to details accurately but also reduce the oversampling in flat regions. 
Specifically, We use SAM~\cite{kirillov2023segment} and a pre-trained classifier~\cite{xie2021segformer} to get segmented regions with fine masks using monocular images as input. Besides, instead of directly using the proportion of each region in the image as sampling weight, we use power functions to remap original proportions, aiming to pay more attention to small details. We define a region-variant weight for ray sampling as follows:
\begin{equation}
W_{i, j} = \frac{(N_i^{j})^{\frac{1}{\delta}}}{\sum{(N_i^{j})^{\frac{1}{\delta}}}},
\end{equation}

\noindent where $N_{i}^j$ is the number of pixels for segmented region $S_j$ of image $I_i$ and $\delta$ is a hyperparameter indicating the importance of detail areas. For each image $I_i$, $q$ sampled rays are assigned by $W_{i,j} * q$ to different segmented regions. The hybrid ray sampling method guided by segmentation prior ensures that detailed objects can be sampled in each iteration, which is beneficial for reconstructing details.

% 在每次迭代，图像I的q个采样射线根据权重被分配到不同区域

\noindent\textbf{Weight-based Point Importance Sampling}\label{ssec:Points-Resampling}
% 首先简述一下HVS，引出原始PDF的问题：分段常数使采样效率低 -> 引出我们的具体做法
% HVS被广泛的用于基于NeRF的方法。它使用coarse-to-fine的采样策略，通过粗采样阶段得到的体素密度引导精细阶段的采样，使采样点更多的分布于物体表面，即体素密度较大的采样区间内。但是，HVS通过常数函数来建模区间内的概率密度，采样点在区间内是均匀分布的，仍是相对粗略的采样。与？？类似，考虑到单调、简单和陡峭的梯度，我们利用指数函数代替常数函数来插值概率密度函数w(t),使用区间边界的权重，使得在区间内也能基于权重调整采样点的分布。具体的，由于粗采样点是沿射线等间距分布的，为了便于运算，我们可以将单个采样区间[t_i,t_i+1]映射到正则化[0,1]区间。因此，正则化区间内一点s的权重定义为：w^(s)=w(t)=w((t_i+1-t_i-1)s + t_i)，其中t为采样区间中的点。设w^(0)=m,w^(1)=n,正则化区间内的概率密度函数可以表示为公式？？。对于fine阶段采样，我们首先遵循HVS将采样点分配到不同区间内，之后通过逆变换采样得到残余概率r对应的区间内采样位置x，见公式？？。在实际采样中，残余概率r由该点与其所在区间的积分下限的累计概率的差值得到：
% 由公式？？所示，其中W()为沿着整条ray的累计分布函数。
HVS methodology has been widely used in NeRF-based methods, utilizing a coarse-to-fine sampling strategy. The strategy uses the weights of points obtained in the coarse stage to guide the sampling of the fine stage, so that more sampling points are distributed around the surface, i.e., within the interval with larger weight. However, HVS models the PDF in each interval using a constant function, which causes a uniform distribution of points in a single interval, leading to still relatively rough sampling. Similar to \cite{li2023l_0}, considering monotonicity, simplicity, and steep gradient, we use exponential functions instead of constant functions to interpolate PDF through weights at the interval boundaries, so that the distribution of sampling points in the interval can be adjusted. Specifically, since the coarse points are equally spaced along the ray, we can map the single interval $[t_i, t_{i+1}]$ to the normalized $[0, 1]$ interval. Therefore, the weight of point $s$ in the normalized interval is defined as: $\Hat{\omega}(s)=\omega(t)=\omega((t_{i+1}-t_i)s+t_i)$, where $t$ is the points in the unnormalized interval. Assuming $\Hat{\omega}(0)=m, \Hat{\omega}(1)=n$, the PDF of the normalized interval can be expressed as follows:

\begin{equation}
    \Hat{\omega}(s) = m (\frac{n}{m})^s.
\end{equation}

\noindent For fine stage sampling, we first follow HVS to allocate points to different intervals, and then use inverse sampling to obtain the normalized specific position $z$ in the interval corresponding to the residual probability $\Delta{r}$:

\begin{equation}
    \Delta{r} = \int_{0}^{z} m(\frac{n}{m})^s {\rm d}s \Rightarrow z=\frac{\ln \frac{\Delta{r}(\ln n-\ln m)}{m}+1}{\ln n - \ln m}.
\end{equation}

\noindent In actual sampling, the residual probability $\Delta r$ is obtained by the cumulative probability $P_T(\cdot)$ difference between point $t$ and the low limit of integral $t_i$ of interval in which $t$ is located:

\begin{equation}
    \Delta{r} = P_T(t)-P_T(t_i).
\end{equation}

\begin{table}[b]
\centering
% \vspace{-0.1cm}
\begin{tabular}{ll}
\toprule[1pt]
\multicolumn{2}{c}{3D Metrics} \\
\midrule
Acc. & $\mbox{mean}_{c \in C}(\min_{c^*\in C^*}||c-c^*||)$ \\
Comp. & $\mbox{mean}_{c^* \in C^*}(\min_{c\in C}||c-c^*||)$ \\
Chamfer & $\frac{\text{Acc}+\text{Comp}}{2}$\\
Prec. & $\mbox{mean}_{c \in C}(\min_{c^*\in C^*}||c-c^*||<.05)$ \\
Recall & $\mbox{mean}_{c^* \in C^*}(\min_{c\in C}||c-c^*||<.05)$ \\
F-score & $\frac{ 2 \times \text{Perc} \times \text{Recall} }{\text{Perc} + \text{Recall}}$ \\
\bottomrule[1pt]
\end{tabular}
% \vspace{-0.1cm}
\caption{Definitions of 3D metrics: $c$ and $c^*$ are the predicted and ground truth point clouds.
}
\label{tab:metric_defs}
\end{table}
%%%%%%%%%%%%%%%%%%%%%%%%%%%%%%%%%%%%%%%%%%%%%%%%%%%%%%%%%%%%%%%%%%%%%%%%

%%%%%%%%%%%%%%%%%%%%%%%%%%%%%%%%%%%%%%%%%%%%%%%%%%%%%%%%%%%%%%%%%%%%%%%%
\begin{table}[ht]
\centering
% \vspace{-0.1cm}
\begin{tabular}{ll}
\toprule[1pt]
\multicolumn{2}{c}{2D Metrics} \\
\midrule
Abs Rel & $\frac{1}{n}\sum|d-d^{*}|/d^{*}$ \\
Sq Rel & $\frac{1}{n}\sum|d-d^{*}|^2/d^{*}$ \\
RMSE & $\sqrt{\frac{1}{n}\sum|d-d^{*}|^2}$ \\
RMSE log & $\sqrt{\frac{1}{n}\sum|\log(d)-\log(d^{*})|^2}$ \\
$\delta < 1.25^3$ & $\frac{1}{n}\sum(\max(\frac{d}{d^*}, \frac{d^*}{d})<1.25^3)$ \\
\bottomrule[1pt]
\end{tabular}
\vspace{-0.1cm}
\caption{Definitions of 2D depth metrics: $d$ and $d^*$ are the predicted and ground truth depths (the predicted depth is obtained by rendering the predicted mesh).}
\label{tab:metric_defs_2d}
\end{table}
%%%%%%%%%%%%%%%%%%%%%%%%%%%%%%%%%%%%%%%%%%%%%%%%%%%%%%%%%%%%%%%%%%%%%%%%

%%%%%%%%%%%%%%%%%%%%%%%%%%%%%%%%%%%%%%%%%%%%%%%%%%%
\begin{table}[t]
    % \vspace{-0.1cm}
    \centering
    \resizebox{0.479\textwidth}{!}{
    % \normalsize
    \begin{tabular}{lcccc>{\columncolor[gray]{0.902}}c}
    \Xhline{3\arrayrulewidth}
    % \toprule
    Method                                 & Acc. $\downarrow$  & Comp. $\downarrow$ & Prec. $\uparrow$         & Recall $\uparrow$     & \textbf{F-score} $\uparrow$           \\ \hline
    COLMAP~\cite{schonberger2016structure}       & 0.062          & 0.090           & 0.640           & 0.569     &0.600              \\ 
    NeuralRecon~\cite{sun2021neuralrecon}  & 0.042          & 0.090          & 0.747           & 0.574     &0.648              \\ 
    NeRF~\cite{mildenhall2021nerf}            & 0.160           & 0.065           & 0.378           & 0.576     &0.454              \\ 
    NeuS~\cite{wang2021neus}         & 0.105           & 0.124           & 0.448           & 0.378     &0.409              \\
    % \hline
    Manhattan-SDF~\cite{Guo_2022_CVPR}  & 0.052          & 0.072           & 0.709           & 0.587     &0.641              \\                                               
    MonoSDF~\cite{yu2022monosdf}         & 0.048          & 0.068           & 0.673           & 0.558     &0.609              \\ 
    NeuRIS~\cite{wang2022neuris}           & 0.053           & 0.053           & 0.717           & 0.662     &0.688              \\
    HelixSurf~\cite{liang2023helixsurf}           & 0.063           & 0.134           & 0.657           & 0.504     &0.567              \\
    % \hline
    % Base         & 0.047           & 0.051           & 0.756           & 0.686     &0.719              \\
    % Model-A         & 0.040           & 0.043           & 0.800           & 0.734     &0.765              \\
    % Model-B           & 0.038           & 0.043           & 0.821           & 0.746     &0.781              \\
    % Model-C           & 0.039           & 0.044           & 0.828           & 0.754     &0.789              \\
    \textbf{Ours}     &  \textbf{0.038}        & \textbf{0.043}           & \textbf{0.831}      & \textbf{0.761}     & \textbf{0.794}              \\
    \Xhline{3\arrayrulewidth}
    % \bottomrule
    \end{tabular}
    }
    \vspace{-0.3cm}
    \caption{Quantitative results of reconstruction with existing methods over 8 scenes using 3D geometry metrics.}
    \label{tab:scannet-3d}
    % \vspace{-0.4cm}
\end{table}
%%%%%%%%%%%%%%%%%%%%%%%%%%%%%%%%%%%%%%%%%%%%%%%%%%%%%%%%%%%%%%%%%%%%%%%%

%%%%%%%%%%%%%%%%%%%%%%%%%%%%%%%%%%%%%%%%%%%%%%%%%%%
\begin{table}[!ht]
    % \vspace{-0.1cm}
    \centering
    \resizebox{0.479\textwidth}{!}{
    % \normalsize
    % \begin{tabular}{lcccc>{\columncolor[gray]{0.902}}c}
    \begin{tabular}{lccccc}
    \Xhline{3\arrayrulewidth}
    % \toprule
    Method                                 & Abs Rel $\downarrow$  & SQ Rel $\downarrow$ & RMSE $\downarrow$         & RM Log $\downarrow$     & $\delta < 1.25^3$ $\uparrow$           \\ \hline
    COLMAP~\cite{schonberger2016structure}       & 0.125          & 0.096           & 0.383           & 0.254     &0.950              \\ 
    NeuralRecon~\cite{sun2021neuralrecon}  & 0.099          & 0.114          & 0.376           & 0.442     &0.952              \\ 
    NeRF~\cite{mildenhall2021nerf}            & 0.166           & 0.191           & 0.561           & 0.794     &0.900              \\ 
    NeuS~\cite{wang2021neus}         & 0.114           & 0.078           & 0.328           & 0.295     &0.968              \\
    % \hline
    Manhattan-SDF~\cite{Guo_2022_CVPR}  & 0.063          & 0.043           & 0.233           & 0.230     &0.986              \\                                   
    MonoSDF~\cite{yu2022monosdf}           & 0.055           & 0.022           & 0.156           & \textbf{0.094}     & \textbf{0.996}              \\
    NeuRIS~\cite{wang2022neuris}           & 0.051           & 0.025           & 0.170           & 0.117     & 0.991              \\
    HelixSurf~\cite{liang2023helixsurf}           & 0.070           & 0.034           & 0.216           & 0.126     & 0.987              \\
    \textbf{Ours}     &  \textbf{0.042}        & \textbf{0.022}           & \textbf{0.152}      & 0.102     & 0.994              \\
    \Xhline{3\arrayrulewidth}
    % \bottomrule
    \end{tabular}
    }
    \vspace{-0.3cm}
    \caption{Quantitative results of reconstruction with existing methods over 8 scenes using 2D depth metrics.}
    \label{tab:scannet-2d}
    \vspace{-0.4cm}
\end{table}
%%%%%%%%%%%%%%%%%%%%%%%%%%%%%%%%%%%%%%%%%%%%%%%%%%%%%%%%%%%%%%%%%%%%%%%%

% 承接一下上面主要是关注采样过程，这一节主要利用multi-view的一致性加强监督（一方面是细节区域有丰富的视觉特征 -> 增加了多视角特征的一致性监督；另一方面是sharp shape区域的normal不准 -> 利用多视角法线一致性做过滤和引导重要性采样）
\subsection{Multi-view Consistency}\label{sec:Multi-view Consistency}
The multi-level importance sampling strategy in Sec. \ref{sec:Multi-level-Resampling} ensures more attention to detailed regions and improves sampling efficiency. To further improve the reconstruction of details, we utilize multi-view consistency to enhance supervision in the training phase. 

\noindent\textbf{Multi-view Feature Consistency}
% 首先阐述多视角一致性在MVS的方法中很常见 -> 引出现有一些利用多视角光度的方法 -> 我们利用更鲁棒的feature的一致性
Guiding geometry reconstruction with multi-view consistency is popular in MVS and recent neural surface reconstructions. Based on the observation that detailed regions mostly have sharp shapes or varied textures, we introduce multi-view consistency constraints to enhance the learning of these regions with rich visual features. 
Different from \cite{wang2022neuris} using multi-view photometric consistency, we utilize more robust deep image features to perform explicit supervision. Following ~\cite{zhang2021learning}, features are extracted by a pre-trained convolutional neural network (CNN) for supervised MVS. Since we sample points around the surface as many as possible, as described in Sec. \ref{ssec:Points-Resampling}, the distance between the points closest to the surface on both sides is small. So similar to \cite{chen2023recovering}, we use linear interpolation to find the zero-crossing of the SDF values as surface points between the last positive SDF values at $\mathbf{x}(t_i)$ and the first consecutive negative values at $\mathbf{x}(t_{i+1})$, which reduces extra calculation compared with applying ray tracing. 
% To find surface points to perform feature consistency, instead of applying ray tracing, similar to \cite{chen2023recovering}, we use linear interpolation to find the zero-crossing of the SDF values between the last positive SDF values $\mathbf{x}(t_i)$ and the first consecutive negative values $\mathbf{x}(t_{i+1})$. 
After deriving the interpolated surface point $\hat{\mathbf{x}}$, the final multi-view feature consistency loss is formulated as follows:
\begin{equation} \label{multi-view feat loss}
    % \mathcal{L}_{\mathit{feat.}}=\frac{\sum_{\boldsymbol p_j \in \boldsymbol P_i}O_{\boldsymbol p_j} \lVert \hat{\text SDF}(\boldsymbol p_j)-\text SDF(\boldsymbol p_j)\rVert_1}{\sum_{\boldsymbol p_j \in \boldsymbol P_i}O_{\boldsymbol p_j}},
    \mathcal{L}_{\mathit{feat.}}=\frac{1}{N_c N_s}\sum_{i=1}^{N_s}{|\mathbf{F}_0(p_0)-\mathbf{F}_i(\mathbf{K}_i(\mathbf{R}_i\hat{\mathbf{x}}+\mathbf{t}_i))|},
\end{equation}
where $N_c$ and $N_s$ are the numbers of feature channels and neighboring source views respectively, {$\mathbf{F}$} is the extracted feature map for a specific view, $p_0$ is the pixel through which the ray casts in the reference view, and $\{\mathbf{K}_i, \mathbf{R}_i, \mathbf{t}_i\}$ are the parameters of the $i$-th neighboring source view.

\noindent\textbf{Multi-view Normal Consistency}
Inspired by~\cite{yu2022monosdf,wang2022neuris}, we incorporate the normal prior estimated by a pre-trained normal predictor~\cite{bae2021estimating} into the optimization of neural implicit surfaces. However, it often leads to over-smoothed results since inaccurate predictions of normal maps, especially in thin and detailed geometries. \cite{wang2022neuris} evaluates the normal prior using the photometric consistency, resulting in incorrectly filtering out the faithful priors for simple planar regions with rich texture. Instead, we directly utilize the prior uncertainty from multi-view to filter the unreliable priors, based on the assumption that a prior is correct if it is consistent with 
% based on the observation that the inaccurate priors usually have a large variance from multiple views
other views. For pixel $p$, the normal uncertainty is presented as:
\begin{equation}
    \mathbf{u}=\frac{1}{N_s}\sum_{i=1}^{N_s}{\arccos(\frac{\mathcal{N}_0(p_0)\cdot\mathcal{N}_i(\mathbf{K}_i(\mathbf{R}_i\hat{\mathbf{x}}+\mathbf{t}_i))}{\|\mathcal{N}_0(p_0)\|\|\mathcal{N}_i(\mathbf{K}_i(\mathbf{R}_i\hat{\mathbf{x}}+\mathbf{t}_i))\|}}),
\end{equation}
where $\mathcal{N}$ is the normal prior for a specific view. Once obtaining the uncertainty $\mathbf{u}$ for sample pixel $p$, the corresponding training weight of normal prior can be given by the indicator function:

\begin{equation}
\Omega(p)=\left\{
\begin{aligned}
1 \ & \text{if} \ \boldsymbol{\mathbf{u}} \le \tau \\
0 \ & \text{if} \ \boldsymbol{\mathbf{u}} > \tau
\end{aligned}
\right.
,
\end{equation}
\noindent where $\tau$ is a hyperparameter indicating the threshold of average angular difference of normal priors between multiple viewpoints. 
Additionally, we utilize the uncertainty to guide ray importance sampling, by increasing the sampling probabilities for regions with unreliable priors. 
% 增加normal uncertainty高的像素点采样关注度

%%%%%%%%%%%%%%%%%%%%%%%%%%%%%%%%%%%%%%%%%%%%%%%%%%%%%%%
\begin{figure*}[ht]
\vspace{-0.3cm}
\centering
\includegraphics[width=0.96\textwidth]{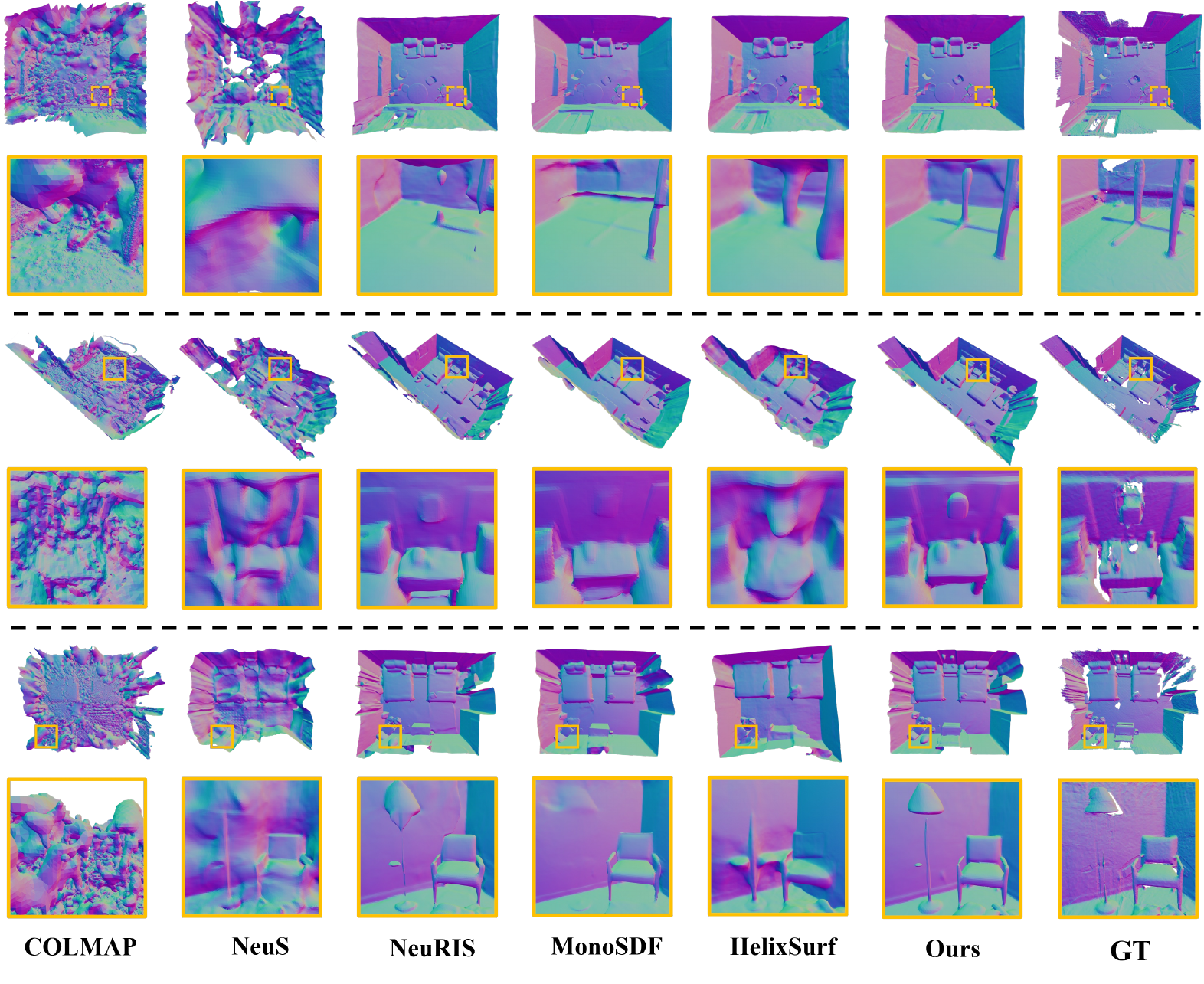}
\vspace{-0.3cm}
\caption{Qualitative results on ScanNet dataset~\cite{dai2017scannet}. For each indoor scene, the first row is the top view of the whole room, and the second row is the details of the masked region. The reconstruction results of FD-NeuS visually have similar scene integrity to those of NeuRIS~\cite{wang2022neuris} and MonoSDF~\cite{yu2022monosdf}. The detailed areas are preserved better than other methods.}
\label{Fig. vis-comp}
\end{figure*}

%%%%%%%%%%%%%%%%%%%%%%%%%%%%%%%%%%%%%%%%%%%%%%%%%%%%%%%

\subsection{Loss Functions}\label{sec:Training}
During training, we sample $\mathit{q}$ pixels per batch, and for each pixel, we sample $\mathit{n}$ points along the corresponding ray. The overall loss can be written as: 

\begin{equation}
\mathcal{L} = \mathit{\lambda}_{1}\mathcal{L}_{\mathit{rgb}} +\mathit{\lambda}_{2}\mathcal{L}_{\mathit{normal}}+\mathit{\lambda}_{3}\mathcal{L}_{\mathit{feat.}} +\mathit{\lambda}_{4}\mathcal{L}_{\mathit{eik}},
\end{equation}

\noindent where $\mathcal{L}_{\mathit{rgb}}$ is the color loss:

\begin{equation}
\mathcal{L}_{\mathit{rgb}} = \frac{1}{\mathit{q}}\sum_{p}\| C(p)-\hat C(p)\|_1,
\end{equation}

\noindent where $C(p)$ and $\hat C(p)$ are ground truth colors and the rendered colors respectively. The normal loss $\mathcal{L}_{\mathit{normal}}$ is denoted by:

\begin{equation}
\mathcal{L}_{\mathit{normal}} = \frac{1}{\mathit{q}}\sum_{p}\| \mathcal{N}(p)-\hat{\mathcal{N}}(p)\|_1\cdot \Omega(p),
\end{equation}

\noindent where $\mathcal{N}(p)$ denotes the predicted monocular normal priors transformed to the world coordinate system and $\hat{\mathcal{N}}(p)$ is the rendered normals. Following~\cite{gropp2020implicit}, the loss $\mathcal{L}_{\mathit{eik}}$ to regularize the gradients of SDF is defined as:

\begin{equation}
\mathcal{L}_{\mathit{eik}} = \frac{1}{\mathit{nq}}\sum_{\mathit{n},\mathit{q}}(\| \nabla f_g(\mathbf{x}_{\mathit{n},\mathit{q}}) \|_2-1)^2.
\end{equation}

\noindent The $\mathit{\lambda }_{1}$, $\mathit{\lambda }_{2}$, $\mathit{\lambda }_{3}$, $\mathit{\lambda }_{4}$ represent the weights of rgb loss, normal loss, feature loss and eikonal loss respectively. 
% We set $\mathit{\lambda }_{1}$, $\mathit{\lambda }_{2}$, $\mathit{\lambda }_{3}$, $\mathit{\lambda }_{4}$ to 1,1,1,0.1, respectively.

\section{Experiments}\label{sec:typestyle}
\subsection{Experimental Setup}
\textbf{Dataset} 
We evaluate the performance of our approach on ScanNet (V2)~\cite{dai2017scannet}. We select 8 scenes with relatively rich details from \cite{wang2022neuris} and \cite{liang2023helixsurf} to conduct our experiments, and all images are resized in 640 $\times$ 480 resolution.

\noindent\textbf{Baselines} 
We compare against: (1) classic MVS method: COLMAP~\cite{schonberger2016structure}, (2) TSDF based method: NeuralRecon~\cite{sun2021neuralrecon}, (3) neural volume rendering methods: NeRF~\cite{mildenhall2021nerf}, NeuS~\cite{wang2021neus}, Manhattan-SDF~\cite{Guo_2022_CVPR}, MonoSDF~\cite{yu2022monosdf}, NeuRIS~\cite{wang2022neuris} and HelixSurf~\cite{liang2023helixsurf}. For COLMAP~\cite{schonberger2016structure}, we use ground truth poses to reconstruct point clouds and then use octree $depth=11$ in the Poisson reconstruction to get the mesh. For NeRF~\cite{mildenhall2021nerf}, we use the level set 20 to extract surfaces by following~\cite{wang2022neuris}.

\noindent\textbf{Metrics} We evaluate our method using 3D geometry metrics and 2D depth metrics, defined in \cref{tab:metric_defs} and \cref{tab:metric_defs_2d}. Among these metrics, F-score is recognized as the most representative metric for geometry quality evaluation.
% We evaluate our method using 3D geometry metrics and 2D depth metrics. Among these metrics, F-score is considered as the most representative metric to evaluate geometry quality. The definitions of these metrics are detailed in \cref{tab:metric_defs} and \cref{tab:metric_defs_2d}.

\noindent\textbf{Implementation Details}
The geometry function $f_g$ is modeled by an MLP with 8 hidden layers and the color function $f_c$ is modeled by an MLP with 4 hidden layers. Positional encoding and initialization of the implicit neural representation are similar to~\cite{wang2022neuris}. We train our model for 80k iterations; sample 512 rays per batch and 64+64 points on each ray. The hyperparameter $\delta$ increases linearly from 1 to 2 in the training process. In addition, we divide the training into three stages. We first train 30k iterations without multi-view consistency strategy. From 30k to 50k iterations, we set feature consistency loss weight $\mathit{\lambda }_{3}$ as 0.5. In the remaining iterations, we increase normal uncertainty. After training, we extract a mesh from the SDF by the Marching Cube algorithm~\cite{lorensen1987marching} with the volume size of $512^{3}$. The other hyperparameters used in the experiment are as follows: $\tau = \pi/9$, $\mathit{\lambda }_{1} = 1$, $\mathit{\lambda }_{2} = 1$, $\mathit{\lambda }_{4} = 0.1$.

\subsection{Results}
\noindent\textbf{Qualitative Results} To show the visualized reconstruction
results of our method, we compare our FD-NeuS with different reconstruction methods, including COLMAP~\cite{schonberger2016structure}, NeuS~\cite{wang2021neus}, NeuRIS~\cite{wang2022neuris}, MonoSDF~\cite{yu2022monosdf}, HelixSurf~\cite{liang2023helixsurf} and the ground truth. As shown in ~\cref{Fig. vis-comp}, our method can produce high-quality results, especially in detailed regions (e.g., desks, chairs and lamps). Compared with the ground truth, our visual result in some areas is even better. 
% Please refer to the supplementary for more qualitative results on ScanNet dataset~\cite{dai2017scannet} compared with more methods.

\noindent\textbf{Quantitative Results}
% As we are mainly concerned about the 3D reconstruction quality, we compare our method with different methods using 3D geometry metrics. 
% The quantitative comparison results on ScanNet~\cite{dai2017scannet} are shown in~\cref{tab:scannet-3d} and~\cref{tab:scannet-2d}. Our method keeps a relatively balance in accuracy and completeness and significantly surpass existing methods in overall performance.
The quantitative comparison results of 3D evaluation and 2D evaluation on ScanNet~\cite{dai2017scannet} are shown in~\cref{tab:scannet-3d} and~\cref{tab:scannet-2d} correspondingly. 
In 3D geometry evaluation, our method significantly outperforms the existing methods in overall metrics, which keeps a balance in accuracy and completeness.
For 2D depth evaluation, our method achieves superior performance among almost all existing methods, except MonoSDF~\cite{yu2022monosdf}, which uses dense depth maps as prior.
% According to ~\cref{tab:scannet-3d} and~\cref{tab:scannet-2d}, our method 
% As shown in~\cref{tab:scannet-2d}, for 2D depth metrics, our method can surpass almost all existing methods, except for MonoSDF~\cite{yu2022monosdf}, which uses dense depth maps as prior. More comprehensive 3D geometry metrics evaluation results are shown in~\cref{tab:scannet-3d}, our method keeps a relatively balance in accuracy and completeness, and significantly surpass existing methods in overall performance. 

\subsection{Ablation Study}
\begin{figure}[t]
% \vspace{-0.6cm}
\centering
\includegraphics[scale=0.6]{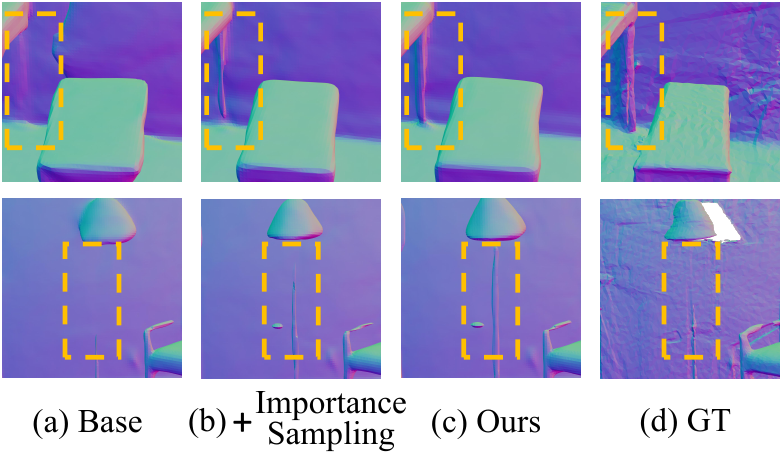}
\vspace{-0.2cm}
\caption{Qualitative results of ablation study. (a) Baseline method. (b) Base model with multi-level importance sampling strategy. (c) Full model. (d) Ground truth.}
\label{Fig. vis-ablation}
\vspace{-0.1cm}
\end{figure}

To validate the effectiveness of our proposed modules, we perform ablation studies on the ScanNet.
% NeuRIS~\cite{wang2022neuris} is adopted as our baseline.
Our base method uses none of our proposed modules, and each module is incrementally added to the baseline to show its efficiency. 
% We train with five settings: (1) \textbf{Base}: without our proposed modules, (2) \textbf{Model-A}: baseline method with RRIS module, (3) \textbf{Model-B}: Model-A with WPIS module, (4) \textbf{Model-C}: Model-B with MVFC, (5) \textbf{Ours}: our full model. 
The corresponding quantitative results are reported in~\cref{tab:ablation}.
According to the results of Base, Model-A, and Model-B, the multi-level importance sampling strategy significantly improves the reconstruction quality by providing more attention to detailed regions and surfaces. The results of Model-C and the full model show that the multi-view consistency provides strengthened supervision, which helps improve the reconstruction accuracy. The ~\cref{Fig. vis-ablation} shows the qualitative results. 

%%%%%%%%%%%%%%%%%%%%%%%%%%%%%%%%%%%%%%%%%%%%%%%%%%%%%%%%%%%%%%%%%%%%%%%%
\begin{table}[t]
    % \vspace{-0.3cm}
    \centering
    \resizebox{0.47\textwidth}{!}{
    \begin{tabular}{lcccccc>{\columncolor[gray]{0.902}}cc}
    \Xhline{3\arrayrulewidth}
     & RRIS & WPIS & MVFC  & MVNU  & Prec. $\uparrow$         & Recall $\uparrow$     & \textbf{F-score} $\uparrow$           \\ \hline
     Base & & &  &   &  0.756 & 0.686 & 0.719 \\
     Model-A    & \ding{51}&  &  &    &  0.800 & 0.734 & 0.765\\
     Model-B     & \ding{51}& \ding{51} & &   & 0.821 & 0.746  & 0.781 \\
     Model-C     & \ding{51}& \ding{51}& \ding{51} &   & 0.828 & 0.754 & 0.789 \\
     Ours    & \ding{51}& \ding{51} & \ding{51}  &  \ding{51} & \textbf{0.831} & \textbf{0.761} & \textbf{0.794} \\
    \Xhline{3\arrayrulewidth}
    \end{tabular}
    }
    \caption{Results of the ablation study on ScanNet dataset. RRIS indicates the Region-based Ray Importance Sampling module. WPIS represents the Weight-based Point Importance Sampling module. MVFC and MVNU indicate Multi-view Feature Consistency and Multi-view Normal Uncertainty correspondingly. 
    % $\downarrow$ indicates the lower score is better and $\uparrow$ indicates the higher score is better.
    }
    \label{tab:ablation}
\end{table}
%%%%%%%%%%%%%%%%%%%%%%%%%%%%%%%%%%%%%%%%%%%%%%%%%%%%%%%%%%%%%%%%%%%%%%%%

%%%%%%%%%%%%%%%%%%%%%%%%%%%%%%%%%%%%%%%%%%%%%%%%%%%%%%%
\linespread{1}
%%%%%%%%%%%%%%%%%%%%%%%%%%%%%%%%%%%%%%%%%%%%%%%%%%%%%%%

\section{Conclusion}\label{sec:Conclusion}
We propose FD-NeuS, a novel neural implicit surface reconstruction method using multi-level importance sampling strategy and multi-view consistency methodology, to recover indoor scenes with fine details. We introduce region-based ray sampling and weight-based point sampling using segmentation prior and piecewise exponential interpolation functions respectively, ensuring more attention on important regions. We additionally use multi-view consistency as supervision and uncertainty to further improve the reconstruction quality of details. Extensive experiments show our method achieves superior performance compared with existing methods on multiple metrics and various scenes. 

\noindent\textbf{Liminations}
Although the training time of our method is greatly improved compared to existing methods, it still takes several hours for each scene, which prevents our method from reconstructing scenes in real-time. Integrating hybrid representations into our model is a promising direction to speed up the training process.

\section{ACKNOWLEDGEMENT}
% This work was funded by the National Natural Science Foundation of China (Grant No. 61972220), Guangdong Natural Science Fund-General Programme Grand No. 2022A1515011234, and University Key Projects stablely funded by Shenzhen Science and Technology Innovation Commission (Grant No. WDZC20200821140447001).
This work was supported by the Shenzhen Science and Technology Major Special Project (KJZD20230923115503007).
\bibliographystyle{IEEEbib}
\bibliography{icip2024template}
% \bibliography{strings,refs}

\end{document}